\documentclass{sty/bmvc2k}
\runninghead{Tan, Le}{MixConv: Mixed Depthwise Convolutional Kernels}

\usepackage{booktabs}     %
\usepackage{multirow}
\usepackage{listings} 
\usepackage{wrapfig}

\title{\Large MixConv: Mixed Depthwise Convolutional Kernels}
\addauthor{Mingxing Tan}{tanmingxing@google.com}{1}
\addauthor{Quoc V. Le}{qvl@google.com}{1}

\addinstitution{
	Google Brain\\
	Mountain View, CA,  USA
}

\begin{document}
\maketitle	
\def\TODO{\textcolor{red}{\emph{TODO: }}}
\newcommand{\TT}[1]{\texttt{#1}}
\newcommand{\BF}[1]{\textbf{#1}}
\newcommand{\IT}[1]{\textit{#1}}

\newcommand\blfootnote[1]{%
	\begingroup
	\renewcommand\thefootnote{}\footnote{#1}%
	\addtocounter{footnote}{-1}%
	\endgroup
}

\newcommand{\km}[1]{{\color{red}[km: #1]}}                                          
\newcommand{\rbg}[1]{{\color{blue}[rbg: #1]}}                                       
\newcommand{\ppd}[1]{{\color{green}[ppd: #1]}}                                      
\newcommand{\bd}[1]{\textbf{#1}}                                                    
\newcommand{\app}{\raise.17ex\hbox{$\scriptstyle\sim$}}                             
\newcommand{\symb}[1]{{\small\texttt{#1}}\xspace}                                   
\newcommand{\mrtwo}[1]{\multirow{2}{*}{#1}}                                         
\def\x{\times}                                                                      
\def\pt{p_\textrm{t}}                                                               
\def\at{\alpha_\textrm{t}}                                                          
\def\xt{x_\textrm{t}}                                                               
\def\CE{\textrm{CE}}                                                                
\def\FL{\textrm{FL}}                                                                
\def\FQ{\textrm{FL}^*}                                                              
\newcommand{\eqnnm}[2]{\begin{equation}\label{eq:#1}#2\end{equation}\ignorespaces}

\newlength\savewidth\newcommand\shline{\noalign{\global\savewidth\arrayrulewidth 
		\global\arrayrulewidth 1pt}\hline\noalign{\global\arrayrulewidth\savewidth}}   
\newcommand{\tablestyle}[2]{\setlength{\tabcolsep}{#1}\renewcommand{\arraystretch}{#2}\centering\footnotesize}
\makeatletter\renewcommand\paragraph{\@startsection{paragraph}{4}{\z@}              
	{.5em \@plus1ex \@minus.2ex}{-.5em}{\normalfont\normalsize\bfseries}}\makeatother
\renewcommand{\dbltopfraction}{1}                                                   
\renewcommand{\bottomfraction}{0}                                                   
\renewcommand{\textfraction}{0}                                                     
\renewcommand{\dblfloatpagefraction}{0.95}                                          
\setcounter{dbltopnumber}{5}

\newcommand{\M}[1]{\mathcal{#1}}

\begin{abstract}
Depthwise convolution is becoming increasingly popular in modern efficient ConvNets, but its kernel size is often overlooked. In this paper, we systematically study the impact of different kernel sizes, and observe that combining the benefits of multiple kernel sizes can lead to better accuracy and efficiency. Based on this observation, we propose a new mixed depthwise convolution  (\BF{MixConv}), which naturally mixes up multiple kernel sizes in a single convolution. As a simple drop-in replacement of vanilla depthwise convolution, our MixConv improves the accuracy and efficiency for existing MobileNets on both ImageNet classification and COCO object detection. To demonstrate the effectiveness of MixConv, we integrate it into AutoML search space and develop a new family of models, named as MixNets, which outperform previous mobile models including MobileNetV2 \cite{mobilenetv218} (ImageNet top-1 accuracy +4.2\%), ShuffleNetV2 \cite{shufflenetv218} (+3.5\%), MnasNet \cite{mnas19} (+1.3\%), ProxylessNAS \cite{proxyless19} (+2.2\%), and FBNet \cite{fbnet19} (+2.0\%).
In particular, our MixNet-L achieves a new state-of-the-art 78.9\% ImageNet top-1  accuracy under typical mobile settings (<600M FLOPS).
Code is at \url{ https://github.com/tensorflow/tpu/tree/master/models/official/mnasnet/mixnet}.

\end{abstract}

\section{Introduction}                                                              
\label{sec:intro} 

Convolutional neural networks (ConvNets) have been widely used in image classification, detection, segmentation, and many other applications. A recent trend in ConvNets design is to improve both accuracy and efficiency. Following this trend, depthwise convolutions are becoming increasingly more popular in modern ConvNets, such as MobileNets \cite{mobilenetv117,mobilenetv218}, ShuffleNets \cite{shufflenet17,shufflenetv218}, NASNet \cite{nas_imagenet18}, AmoebaNet \cite{amoebanets18}, MnasNet \cite{mnas19},  and EfficientNet \cite{efficientnet19}. Unlike regular convolution, depthwise convolutional kernels are applied to each individual channel separately, thus reducing the computational cost by a factor of $C$, where $C$ is the number of channels. While designing ConvNets with depthwise convolutional kernels, an important but often overlooked factor is kernel size. Although conventional practice is to simply use 3x3  kernels  \cite{mobilenetv117,mobilenetv218,shufflenet17,shufflenetv218,xception17,nas_imagenet18}, recent research results have shown larger kernel sizes such as 5x5 kernels \cite{mnas19} and 7x7 kernels \cite{proxyless19} can potentially improve model accuracy and efficiency.

In this paper, we revisit the fundamental question: \emph{do larger kernels always achieve higher accuracy?} Since first observed in AlexNet \cite{alexnet12}, it has been well-known that each convolutional kernel is responsible to capture a local image pattern, which could be edges in early stages  and objects in later stages. Large kernels tend to capture high-resolution patterns with more details at the cost of more parameters and computations, but do they always improve accuracy? To answer this question, we systematically study the impact of kernel sizes based on MobileNets \cite{mobilenetv117,mobilenetv218}. Figure 1 shows the results. As expected, larger kernel sizes significantly increase the model size with more parameters; however, model accuracy first goes up from 3x3 to 7x7, but then drops down quickly when the kernel size is larger than 9x9, suggesting very large kernel sizes can potentially hurt both accuracy and efficiency. In fact, this observation aligns to the very first intuition of ConvNets: in the extreme case that kernel size is equal to the input resolution, a ConvNet simply becomes a fully-connected network, which is known to be inferior [7]. This study suggests the limitations of single kernel size: we need both large kernels to capture high-resolution patterns and small kernels to capture low-resolution patterns for better model accuracy and efficiency.

\begin{figure}[t ]    
	\begin{tabular}{cc}                                                                
		\bmvaHangBox{\includegraphics[width=0.46\linewidth]{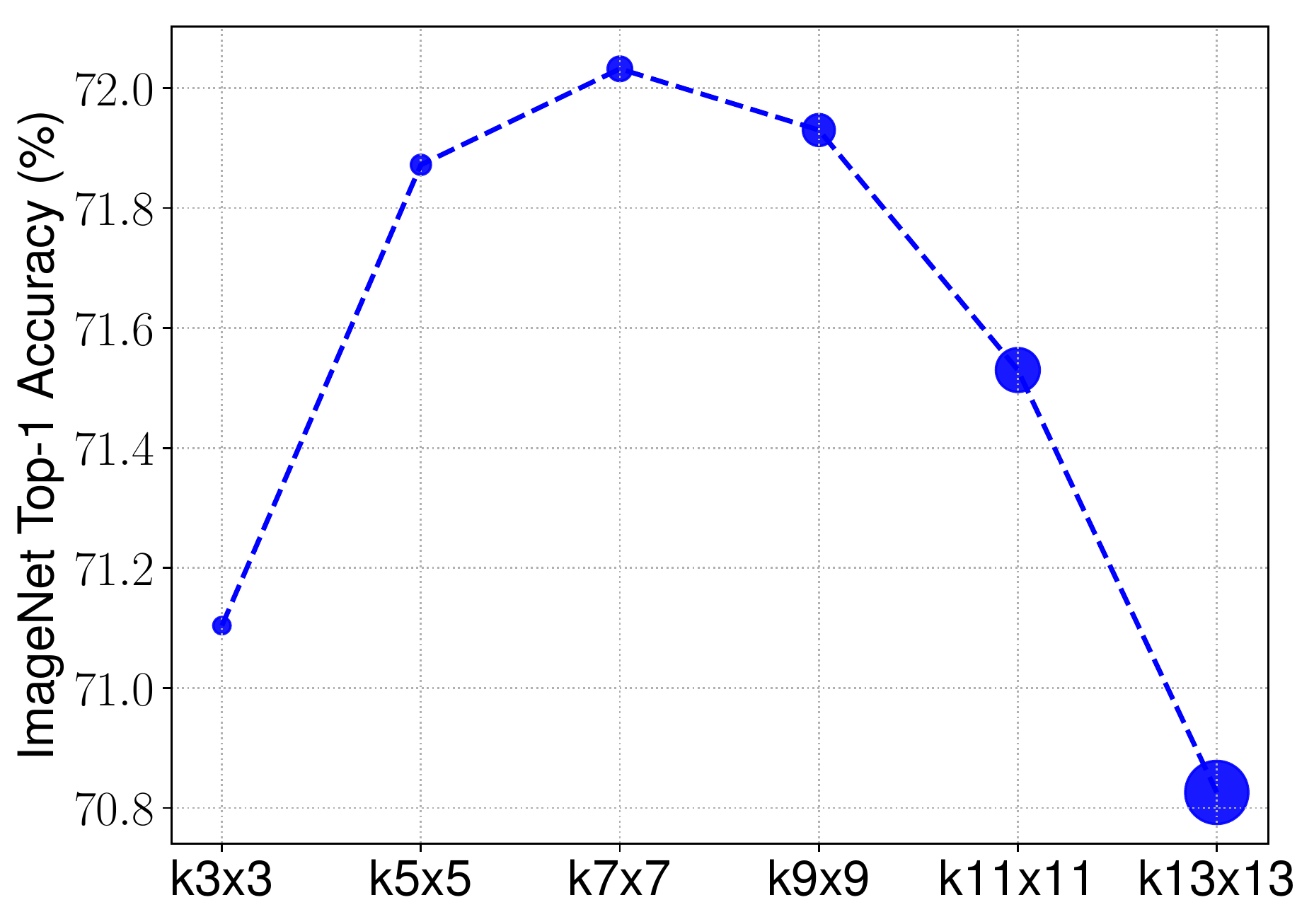}} &
		\bmvaHangBox{\includegraphics[width=0.46\linewidth]{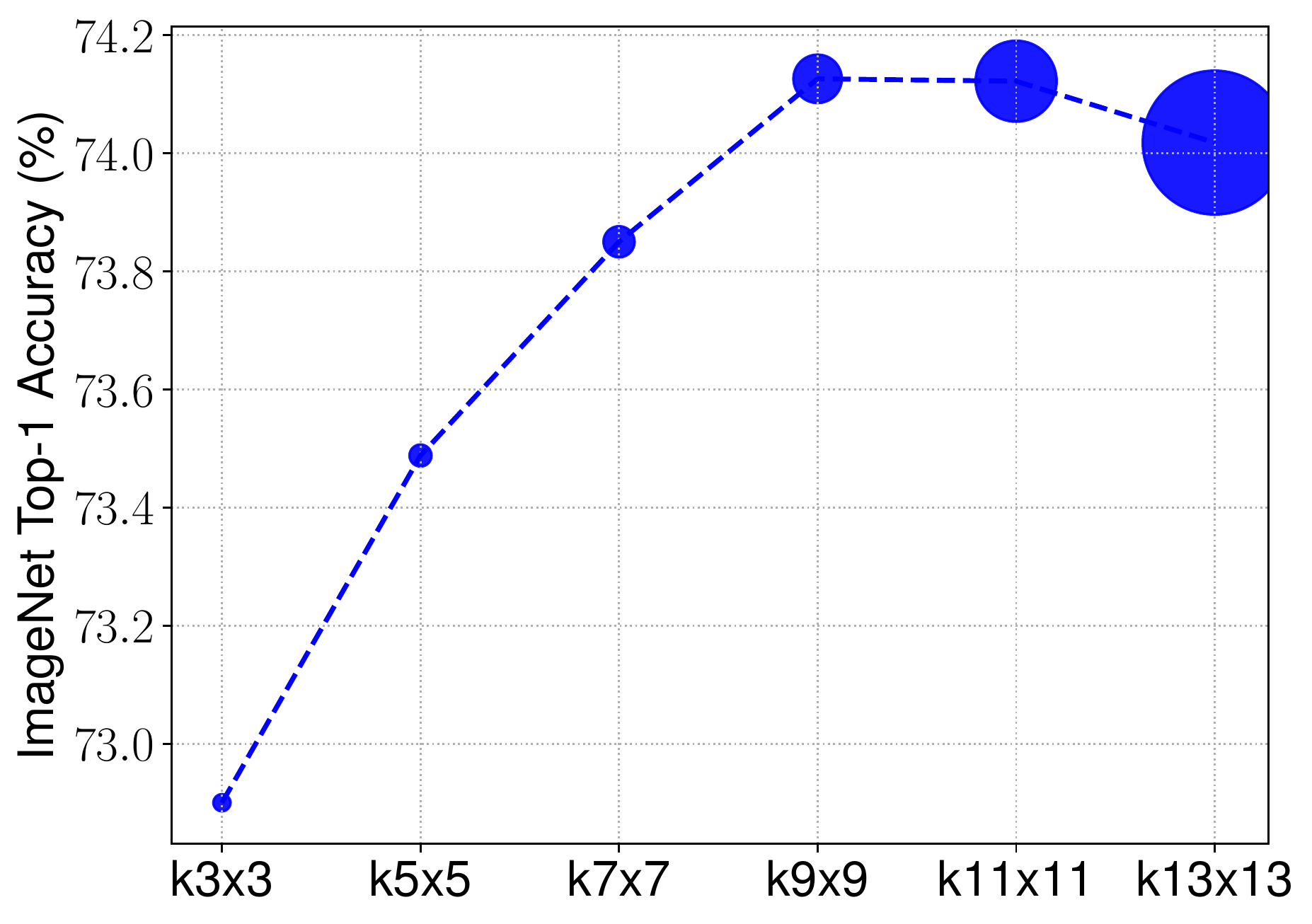}} \\
		(a) MobileNetV1 & (b) MobileNetV2
	\end{tabular}                                                                       
    \caption{
        \BF{Accuracy vs kernel sizes} -- Each point represents a model variant of MobileNet V1\cite{mobilenetv117} and V2 \cite{mobilenetv218}, where model size is represented by point size. Larger kernels lead to more parameters, but the accuracy actually drops down when kernel size is larger than 9x9.
    }                                                                  
	\label{fig:limits}                                                                  
\end{figure}  %
\begin{figure}                                           
	\centering                                                                  
	\includegraphics[width=0.95\linewidth,keepaspectratio=true]{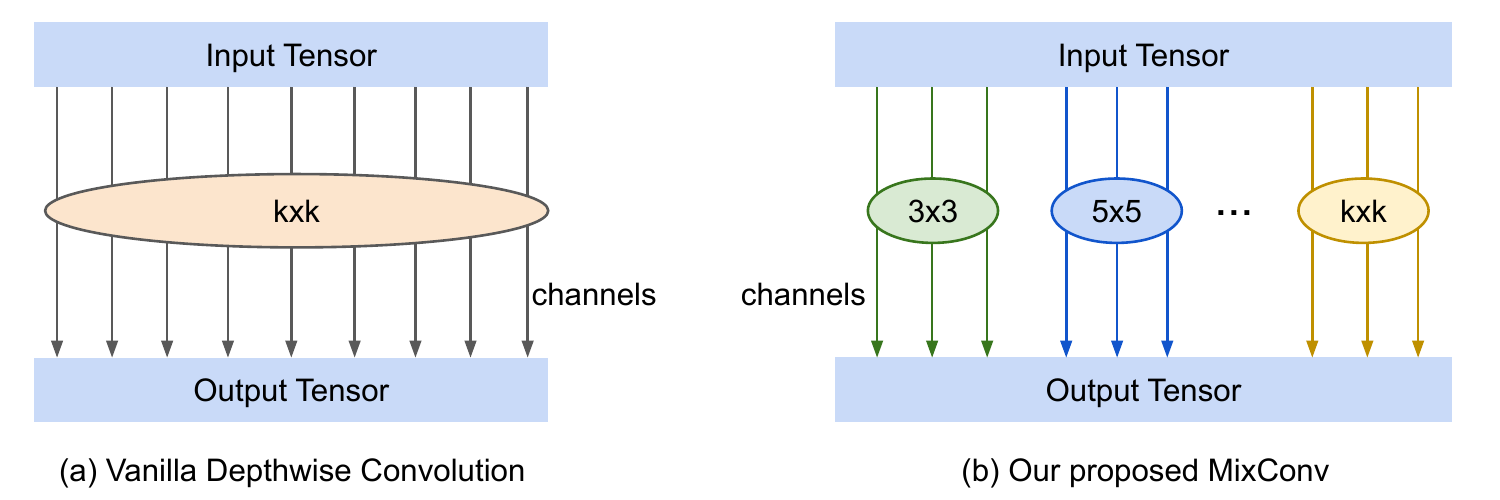}
   \vskip -0.1in
	\caption{
		\textbf{Mixed depthwise convolution (MixConv)} -- Unlike vanilla depthwise convolution that applies a single kernel to all channels, MixConv partitions channels into groups and apply different kernel size to each group.
	}
	\label{fig:mixconv}
\end{figure}
 
Based on this observation, we propose a \emph{mixed depthwise convolution (MixConv)}, which mixes up different kernel sizes in a single convolution op, such that it can easily capture different patterns with various resolutions. Figure 2 shows the structure of MixConv, which partitions channels into multiple groups and apply different kernel sizes to each group of channels. We show that our MixConv is a simple drop-in replacement of vanilla depthwise convolution, but it can significantly improve MobileNets accuracy and efficiency on both ImageNet classification and COCO object detection.

To further demonstrate the effectiveness of our MixConv, we leverage neural architecture search \cite{mnas19} to develop a new family of models named as \emph{MixNets}. Experimental results show our MixNet models significantly outperform all previous mobile ConvNets, such as ShuffleNets \cite{shufflenetv218,shufflenet17}, MnasNet \cite{mnas19}, FBNet \cite{fbnet19}, and ProxylessNAS \cite{proxyless19}. In particular, our large-size MixNet-L achieves a new state-of-the-art 78.9\% ImageNet top-1 accuracy under typical model size and FLOPS settings. %
\section{Related Work}
\label{sec:related}   

\paragraph{Efficient ConvNets:} In recent years, significant efforts have been spent on improving ConvNet efficiency, from more efficient convolutional operations \cite{squeezenet16,mobilenetv117, xception17}, bottleneck layers \cite{mobilenetv218,resnext17}, to more efficient  architectures \cite{mnas19,fbnet19,proxyless19}. In particular, depthwise convolution has been increasingly popular in all mobile-size ConvNets, such as MobileNets \cite{mobilenetv117,mobilenetv218}, ShuffleNets\cite{shufflenet17,shufflenetv218}, MnasNet \cite{mnas19}, and beyond \cite{xception17,nas_imagenet18,amoebanets18}. Recently, EfficientNet \cite{efficientnet19} even achieves both state-of-the-art ImageNet accuracy and ten-fold better efficiency by extensively using depthwise and pointwise convolutions. Unlike regular convolution, depthwise convolution  performs convolutional kernels for each channel separately, thus reducing parameter size and computational cost. Our proposed MixConv generalizes the concept of depthwise convolution, and can be considered as a drop-in replacement of vanilla depthwise convolution. 

\paragraph{Multi-Scale Networks and Features:} Our idea shares a lot of similarities to prior multi-branch ConvNets, such as Inceptions \cite{inceptionv316,inceptionv417}, Inception-ResNet \cite{inceptionresnet17}, ResNeXt \cite{resnext17}, and NASNet \cite{nas_imagenet18}. By using multiple branches in each layer, these ConvNets are able to utilize different operations (such as convolution and pooling) in a single layer. Similarly, there are also many prior work on combining multi-scale feature maps from different layers, such as DenseNet \cite{densenet17,multiscaledense18} and feature pyramid network \cite{fpn17}. However, unlike these prior works that mostly focus on changing the macro-architecture of neural networks in order to utilize different convolutional ops, our work aims to design a drop-in replacement of a single depthwise convolution, with the goal of easily utilizing different kernel sizes without changing the network structure.

\paragraph{Neural Architecture Search:} Recently, neural architecture search \cite{nas_cifar17,nas_imagenet18,pnas18,diffnas18,mnas19} has achieved better performance than hand-crafted models by automating the design process and learning better design choices. Since our MixConv is a flexible operation with many possible design choices,  we employ existing architecture search methods similar to \cite{mnas19,fbnet19,proxyless19} to develop a new family of MixNets by adding our MixConv into the search space.  %
\section{MixConv}
\label{sec:mixconv}

The main idea of MixConv is to mix up multiple kernels with different sizes in a single depthwise convolution op, such that it can easily capture different types of patterns from input images. In this section, we will discuss the feature map  and design choices for MixConv.

\subsection{MixConv Feature Map}
We start from the vanilla depthwise convolution. Let $X^{(h, w, c)}$  denotes the input tensor with shape $(h, w, c)$, where $c$ is the spatial height, $w$ is the spatial width, and $c$ is the channel size. Let  $W^{(k, k, c, m)}$ denotes a depthwise convolutional kernel, where $k \times k$ is the kernel size, $c$ is the input channel size, and $m$ is the channel multiplier.
For simplicity, here we assume kernel width and height are the same $k$, but it is straightforward to generalize to cases where kernel width and height are different.s
The output tensor $Y^{(h, w, c \cdot m)}$ would have the same spatial shape $(h, w)$ and multiplied output channel size $m \cdot c$, with each output feature map value calculated as:

\begin{equation}
Y_{x, y, z} = \sum_{-\frac{k}{2} \le i \le \frac{k}{2}, -\frac{k}{2} \le j \le \frac{k}{2} } {X_{x + i, y + j, z/m} \cdot W_{i, j, z}}, \qquad \forall z=1,...,m\cdot c
\end{equation}

Unlike vanilla depthwise convolution, MixConv partitions channels into groups and applies different kernel sizes to each group, as shown in Figure \ref{fig:mixconv}. More concretely, the input tensor is partitioned into $g$ groups of virtual tensors $<\hat{X}^{(h, w, c_1)}, ..., \hat{X}^{(h, w, c_g)}>$, where all virtual tensors $\hat{X}$ have the same spatial height $h$ and width $w$, and their total channel size is equal to the original input tensor:  $c_1 + c_2 + ... + c_g = c$. Similarly, we also partition the convolutional kernel into $g$ groups of virtual kernels $<\hat{W}^{(k_1, k_1, c_1, m)}, ..., \hat{W}^{(k_g, k_g, c_g, m)}>$. For $t-$th group of virtual input tensor and kernel, the corresponding virtual output is calculated as:

\begin{equation}
\hat{Y}_{x, y, z}^t = \sum_{-\frac{k_t}{2} \le i \le \frac{k_t}{2}, -\frac{k_t}{2} \le j \le \frac{k_t}{2} } {\hat{X}_{x + i, y + j, z/m}^t \cdot \hat{W}_{i, j, z}^t}, \qquad \forall z=1,...,m\cdot c_t
\end{equation}

\noindent The final output tensor is a concatenation of all virtual output tensors $<\hat{Y}_{x, y, z_1}^1, ..., \hat{Y}_{x, y, z_g}^g>$:

\begin{equation}
Y_{x, y, z_o} = Concat\left(\hat{Y}_{x, y, z_1}^1, ..., \hat{Y}_{x, y, z_g}^g\right) 
\end{equation}

\noindent where $z_o = z_1 + ... + z_g = m \cdot c$ is the final output channel size.

\definecolor{codegreen}{rgb}{0,0.5,0}                                            
\definecolor{codeblue}{rgb}{0.25,0.5,0.5}                                        
\definecolor{codegray}{rgb}{0.6,0.6,0.6} 

\lstset{                                                                            
	backgroundcolor=\color{white},                                                    
	basicstyle=\fontsize{7.5pt}{8.5pt}\fontfamily{lmtt}\selectfont,                   
	columns=fullflexible,                                                             
	breaklines=true,                                                                  
	captionpos=b,                                                                     
	commentstyle=\fontsize{8pt}{9pt}\color{codegray},                                 
	keywordstyle=\fontsize{8pt}{9pt}\color{codegreen},                                
	stringstyle=\fontsize{8pt}{9pt}\color{codeblue},                                  
	frame=tb,                                                                         
	otherkeywords ={self},
} 

\begin{wrapfigure}{r}{0.61\linewidth}
\tiny                                                                                          
\begin{lstlisting}[language=python]   
def mixconv(x, filters, **args):                                       
    # x: input features with shape [N,H,W,C]                                        
    # filters: a list of filters with shape [K_i, K_i, C_i, M_i] for i-th group.
    G = len(filters)    # number of groups.
    y = []                                                                     
    for xi, fi in zip(tf.split(x, G, axis=-1), filters):
        y.append(tf.nn.depthwise_conv2d(xi, fi, **args))
    return tf.concat(y, axis=-1)
\end{lstlisting}       
\vskip -0.1in                                                             
\caption{A demo of TensorFlow MixConv.}                            
\label{fig:code}                                                                    
\end{wrapfigure} 
 Figure \ref{fig:code} shows a simple demo of TensorFlow python implementation for MixConv.  On certain platforms, MixConv could be implemented as a single op and optimized with group convolution. Nevertheless, as shown in the figure, MixConv can be considered as a  simple drop-in replacement of vanilla depthwise convolution.

\subsection{MixConv Design Choices}

MixConv is a flexible convolutional op with several design choices:

\paragraph{Group Size $g$:} It determines how many different types of kernels to use for a single input tensor. In the extreme case of $g=1$, a MixConv becomes equivalent to a vanilla depthwise convolution. In our experiments, we find $g=4$ is generally a safe choice for MobileNets, but with the help of neural architecture search, we find it can further benefit the model efficiency and accuracy with a variety of group sizes from 1 to 5.

\paragraph{Kernel Size Per Group:} In theory, each group can have arbitrary kernel size. However, if two groups have the same kernel size, then it is equivalent to merge these two groups into a single group, so we restrict each group has different kernel size. Furthermore, since small kernel sizes generally have less parameters and FLOPS, we restrict kernel size always starts from 3x3, and monotonically increases by 2 per group. In other words, group $i$ always has kernel size $2i + 1$. For example, a 4-group MixConv always uses kernel sizes \{3x3, 5x5, 7x7, 9x9\}. With this restriction, the kernel size for each group is predefined for any group size $g$, thus simplifying our design process.

\paragraph{Channel Size Per Group:} In this paper, we mainly consider two channel partition methods: (1) Equal partition: each group will have the same number of filters; (2) Exponential partition: the $i$-th group will have about $2^{-i}$ portion of total channels. For example, given a 4-group MixConv with total filter size 32, the equal partition will divide the channels into (8, 8, 8, 8), while the exponential partition will divide the channels into (16, 8, 4, 4).

\paragraph{Dilated Convolution:} Since large kernels need more parameters and computations, an alternative is to use dilated convolution \cite{dilat4edconv16}, which can increase receptive field without extra parameters and computations. However, as shown in our ablation study in Section \ref{subsec:ablation},  dilated convolutions usually have inferior accuracy than large kernel sizes.

\begin{figure}                                                                      
	\begin{tabular}{cc}                                                                
		\bmvaHangBox{\includegraphics[width=0.475\linewidth]{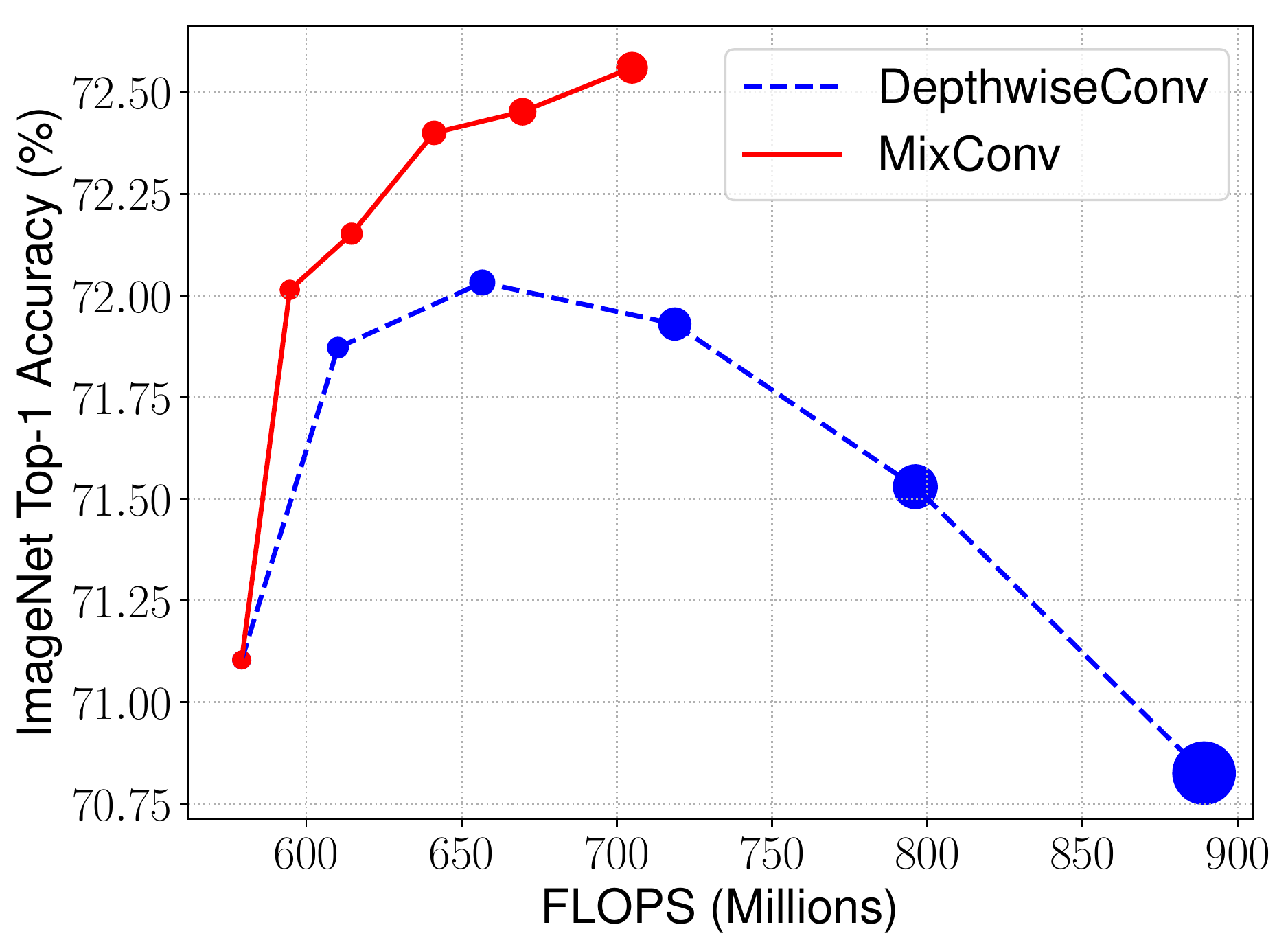}} &
		\bmvaHangBox{\includegraphics[width=0.46\linewidth]{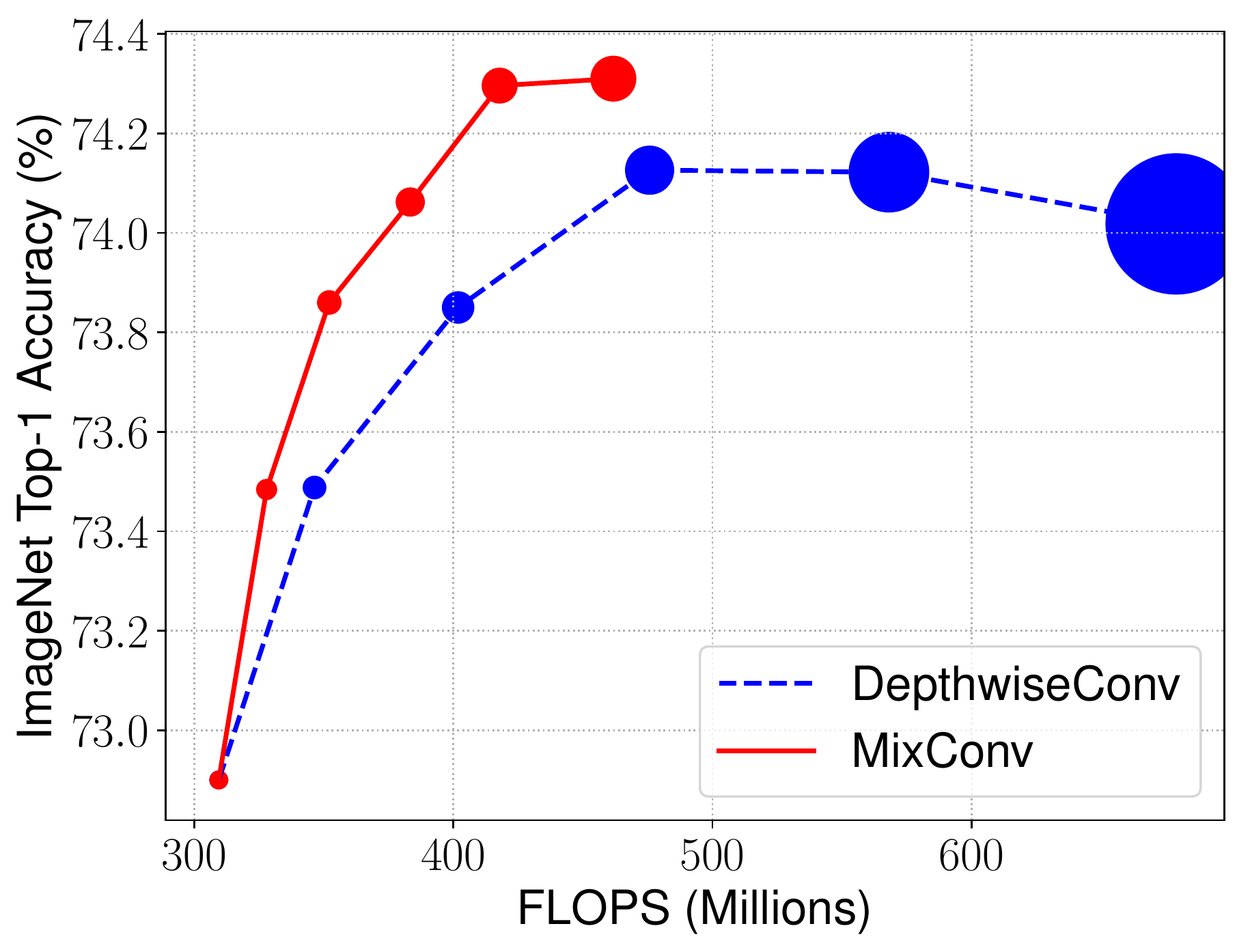}} \\
		(a) MobileNetV1 & (b) MobileNetV2
	\end{tabular}                                                                       
	\caption{
		\textbf{MixConv performance on ImageNet} -- Each point denotes a model with kernel size from 3x3 to 13x13, same as Figure \ref{fig:limits}.  \TT{MixConv} is smaller, faster, and achieves higher accuracy than vanilla depthwise convolutions.
	}                                                         
	\label{fig:mvperf}                                                                  
\end{figure}  

\subsection{MixConv Performance on MobileNets}

Since MixConv is a simple drop-in replacement of vanilla depthwise convolution, we evaluate its performance on classification and detection tasks  with existing MobileNets \cite{mobilenetv117,mobilenetv218}.

\paragraph{ImageNet Classification Performance:} 
Figure  \ref{fig:mvperf} shows the performance of MixConv on ImageNet classification \cite{imagenet15}. Based on MobileNet V1 and V2, we replace all original 3x3  depthwise convolutional kernels with larger kernels or MixConv kernels. Notably, MixConv always starts with 3x3 kernel size and then monotonically increases by 2 per group, so the rightmost point for \TT{MixConv} in the figure has six groups of filters with kernel size \{3x3, 5x5, 7x7, 9x9, 11x11, 13x13\}. In this figure, we observe: (1) MixConv generally uses much less parameters and FLOPS, but its accuracy is similar or better than vanilla depthwise convolution, suggesting mixing different kernels can improve both efficiency and accuracy;  (2) In contrast to vanilla depthwise convolution that suffers from accuracy degradation with larger kernels, as shown in Figure \ref{fig:limits}, MixConv is much less sensitive to very large kernels, suggesting mixing different kernels can achieve more stable accuracy for large kernel sizes.

\paragraph{COCO Detection Performance:} We have also evaluated our MixConv on COCO object detection based on MobileNets. Table \ref{tab:ssd} shows the performance comparison, where our MixConv consistently achieves better efficiency and accuracy than vanilla depthwise convolution.  In particular, compared to the vanilla depthwise7x7, our MixConv357 (with 3 groups of kernels \{3x3, 5x5, 7x7\}) achieves 0.6\% higher mAP on MobileNetV1 and 1.1\% higher mAP  on MobileNetV2 using fewer parameters and FLOPS.

\begin{table*}                                                                   
  \centering                                                                     
  \resizebox{0.8\textwidth}{!}{                                                 
      \begin{tabular}{c|ccc|ccc}                                               
          \toprule[0.2em]        
           & \multicolumn{3}{c}{MobileNetV1~\cite{mobilenetv117}} &  \multicolumn{3}{c}{MobileNetV2~\cite{mobilenetv218}} \\
          Network                           &  \#Params   & \#FLOPS & $mAP$  &  \#Params   & \#FLOPS & $mAP$   \\
          \midrule[0.05em]                                                        
          baseline3x3   & 5.12M  & 1.31B  & 21.7  & 4.35M & 0.79B  & 21.5 \\
          \midrule[0.05em]
          depthwise5x5    & 5.20M  & 1.38B  & 22.3  & 4.47M & 0.87B  & 22.1 \\
          \bf mixconv 35 (ours)&  \bf 5.16M  &  \bf  1.35B  &  \bf 22.2  &  \bf 4.41M &  \bf 0.83B  &  \bf 22.1 \\
          \midrule[0.05em]
          depthwise7x7    & 5.32M  & 1.47B  & 21.8  & 4.64M & 0.98B  & 21.2 \\  
         \bf mixconv 357 (ours)&  \bf  5.22M  &  \bf 1.39B  &  \bf 22.4  &  \bf 4.49M &  \bf 0.88B  &  \bf 22.3 \\
          \bottomrule[0.2em]                                         
      \end{tabular}                                                              
  }
  \vskip 0.1in
  \caption{                                                                      
      \textbf{Performance comparison on COCO object detection.} %
  }
  \label{tab:ssd}
\end{table*}   
 
\subsection{Ablation Study}
\label{subsec:ablation}

To better understand MixConv, we provide a few ablation studies:

\paragraph{MixConv for Single Layer:} 

\begin{figure}[t]
	\centering
	\includegraphics[width=0.7\linewidth,keepaspectratio=true]{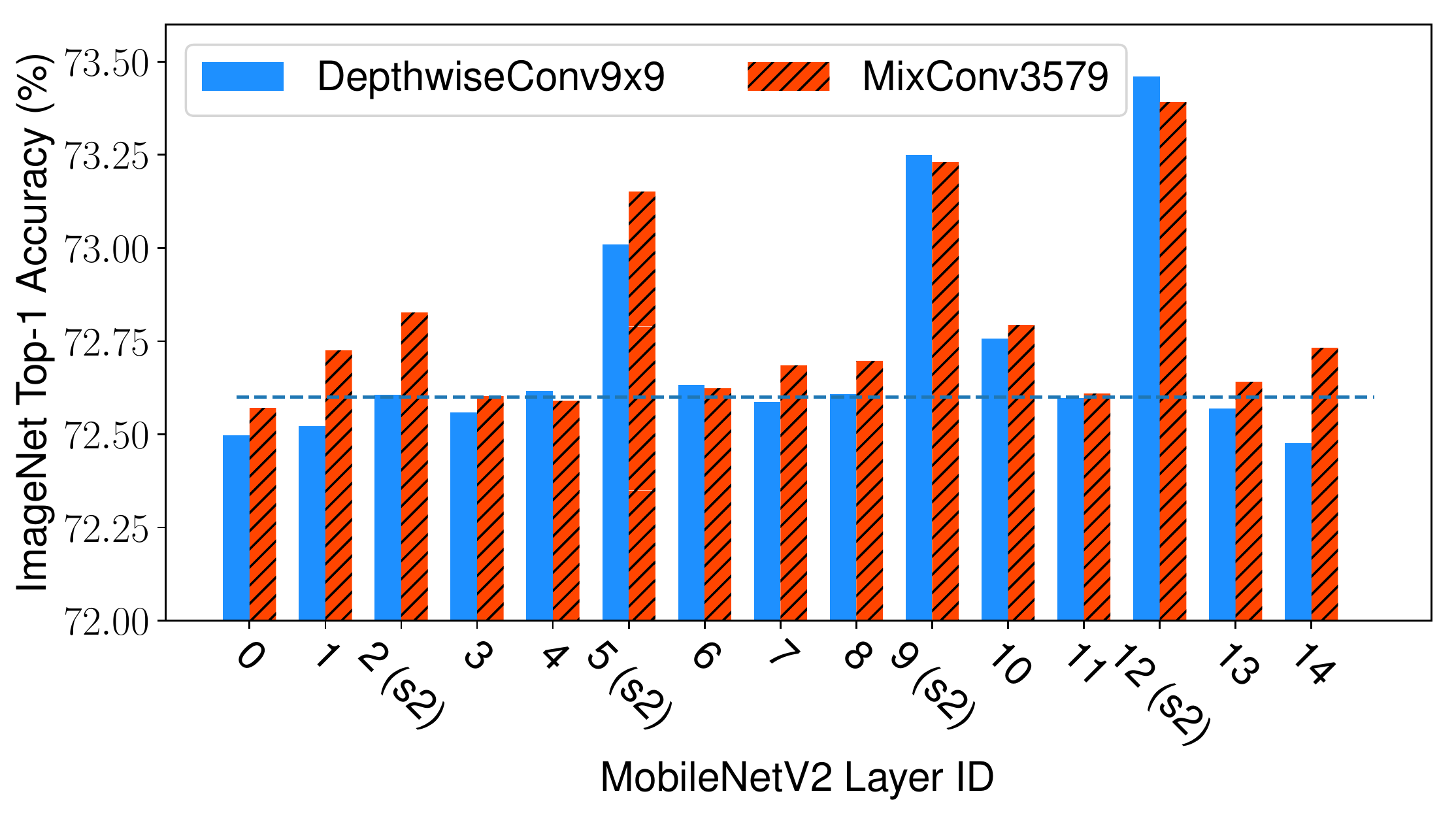}
	\vskip -0.1in
	\caption{
			\textbf{Per-layer impact of kernel size} --  \TT{s2} denotes stride 2, while  others have stride 1. 
	} \label{fig:singlelayer}
\end{figure}

In addition of applying MixConv to the whole network, Figure \ref{fig:singlelayer} shows the per-layer performance on MobileNetV2. We replace one of the 15 layers with either (1) vanilla \TT{DepthwiseConv9x9} with kernel size 9x9; or (2) \TT{MixConv3579} with 4 groups of kernels: \{3x3, 5x5, 7x7, 9x9\}. As shown in the figure, large kernel size has different impact on different layers: for most of layers, the accuracy doesn't change much, but for certain layers with stride 2, a larger kernel can significantly improve the accuracy. Notably, although \TT{MixConv3579} uses only half parameters and FLOPS than the vanilla \TT{DepthwiseConv9x9}, our MixConv achieves similar or slightly better performance for most of the layers.

\begin{figure}                                                                      
	\begin{tabular}{cc}                                                                
		\bmvaHangBox{\includegraphics[width=0.46\linewidth,keepaspectratio=true]{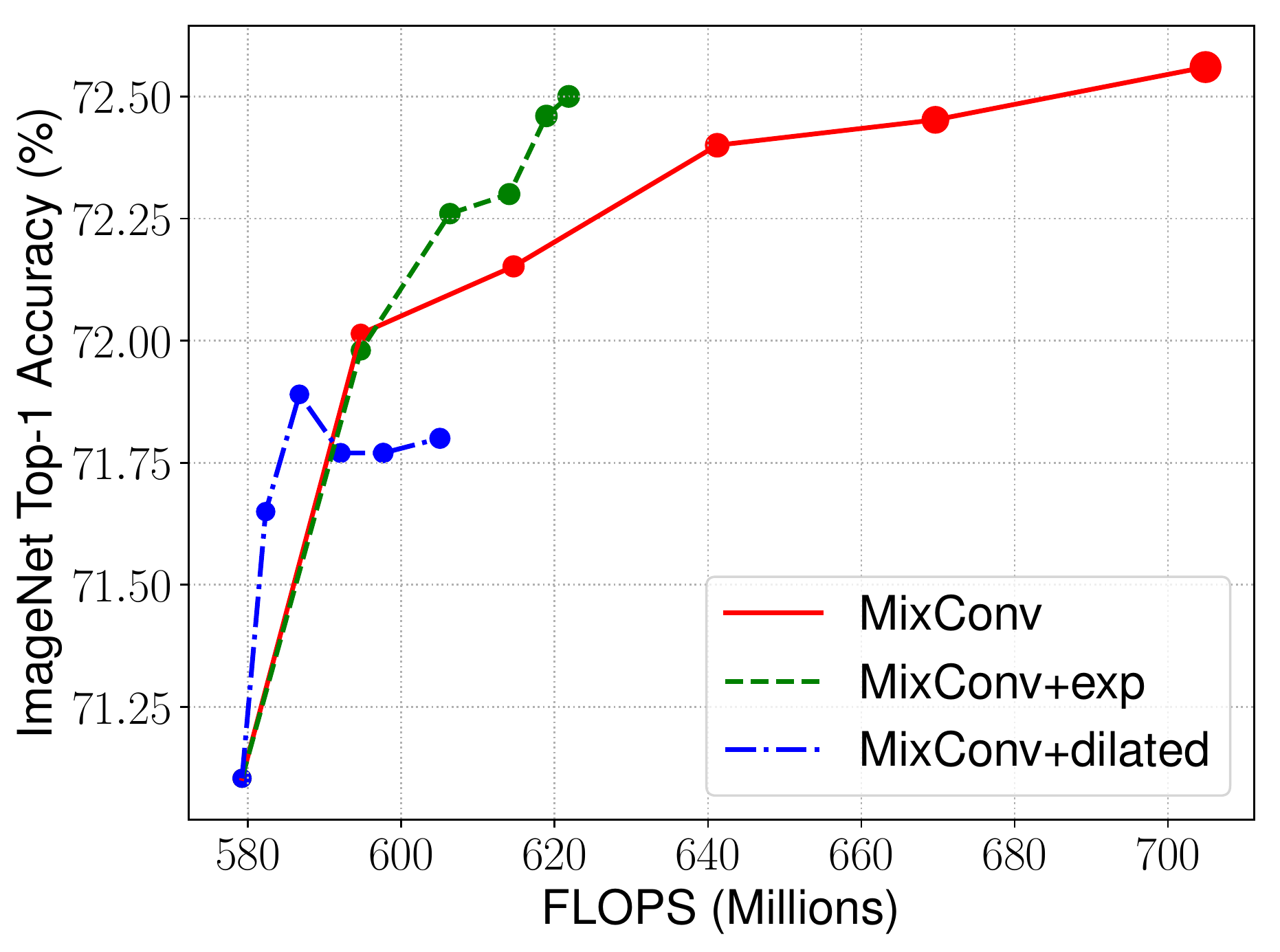}} &
		\bmvaHangBox{\includegraphics[width=0.46\linewidth,keepaspectratio=true]{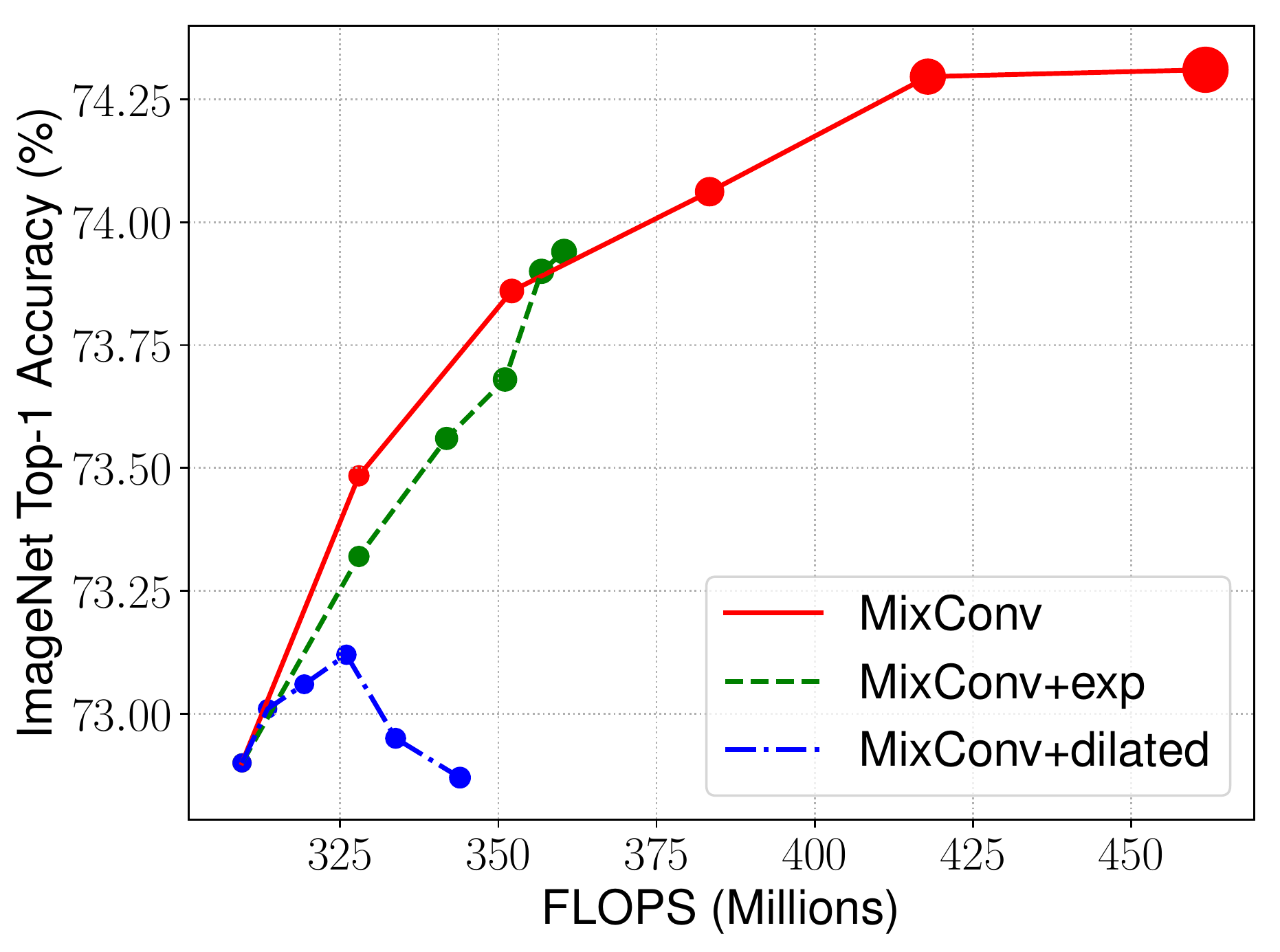}} \\
		(a) MobileNetV1 & (b) MobileNetV2
	\end{tabular}                                                                       
    \caption{
	\BF{Performance of exponential partition (+exp) and dilated kernels (+dilated).} 
    }                                              
	\label{fig:expdilate}                                                                  
\end{figure}  
\paragraph{Channel Partition Methods:} 

Figure \ref{fig:expdilate} compares the two channel partition methods: equal partition (MixConv) and exponential partition (MixConv+exp). As expected, exponential partition requires less parameters and FLOPS for the same kernel size, by assigning more channels to smaller kernels. Our empirical study shows exponential channel partition only performs slightly better than equal partition on MobileNetV1, but there is no clear winner if considering both MobileNet V1 and V2.  A possible limitation of exponential partition is that large kernels won't have enough channels to capture high-resolution patterns.

\paragraph{Dilated Convolution:} 

Figure \ref{fig:expdilate} also compares the performance of dilated convolution (denoted as MixConv+dilated). For kernel size KxK, it uses a 3x3 kernel with dilation rate $(K-1)/2$: for example, a 9x9 kernel will be replaced by a 3x3 kernel with dilation rate 4. Notably, since Tensorflow dilated convolution is not compatible with stride 2, we only use dilated convolutions for a layer if its stride is 1.  As shown in the figure, dilated convolution has reasonable performance for small kernels, but the accuracy drops quickly for large kernels. Our hypothesis is that when dilation rate is big for large kernels, a dilated convolution will skip a lot of local information, which would hurt the accuracy. %
\section{MixNet}
\label{sec:nas}

To further demonstrate the effectiveness of MixConv, we leverage recent progress in neural architecture search to develop a new family of MixConv-based models, named as MixNets.

\subsection{Architecture Search}

Our neural architecture search settings are similar to recent MnasNet \cite{mnas19} and FBNet \cite{fbnet19}, which use MobileNetV2 \cite{mobilenetv218} as the baseline network structure, and search for the best kernel size, expansion ratio, channel size, and other design choices.  Our search space also includes swish activation \cite{swish18,swishsil18}, squeeze-and-excitation module \cite{senet18} (similar to \cite{mnas19}), and grouped convolutions with group size 1 or 2 for 1x1 convolutions (similar to \cite{fbnet19}).
However, unlike these prior works that use vanilla depthwise  convolution as the basic convolutional op, we adopt our proposed MixConv as the search options. Specifically, we have five MixConv candidates with group size $g=1,...,5$:

\begin{table}                                                                      
    \centering                                                                      
    \resizebox{0.95\textwidth}{!}{                                                  
        \begin{tabular}{l|c|cc|cc}                                                
        \toprule [0.2em]                                                            
        Model & Type &  \#Parameters & \#FLOPS & Top-1 (\%) & Top-5 (\%)  \\
        \midrule[0.1em]                 
        MobileNetV1  \cite{mobilenetv117}     & manual  & 4.2M  & 575M & 70.6 & 89.5 \\        MobileNetV2  \cite{mobilenetv218}     & manual  & 3.4M  & 300M & 72.0 & 91.0 \\  
        MobileNetV2 (1.4x)    & manual  & 6.9M  & 585M & 74.7 & 92.5 \\  
        ShuffleNetV2  \cite{shufflenetv218}   & manual  & -  & 299M & 72.6 & - \\
        ShuffleNetV2 (2x)   & manual  & -  & 597M & 75.4 & - \\    
        ResNet-153 \cite{resnet16}    & manual  & 60M  & 11B & 77.0 & 93.3  \\  
        \midrule[0.02em]                                                            
        NASNet-A      \cite{nas_imagenet18}   & auto    & 5.3M  & 564M & 74.0 & 91.3  \\ 
        DARTS  \cite{diffnas18}			  &  auto    & 4.9M & 595M & 73.1 & 91 \\
        MnasNet-A1 \cite{mnas19}      &  auto    & 3.9M  & 312M & 75.2 & 92.5  \\
        MnasNet-A2          &  auto    & 4.8M  & 340M & 75.6 & 92.7 \\
        FBNet-A \cite{fbnet19}	&  auto    & 4.3M  & 249M & 73.0 & - \\
        FBNet-C  	&  auto    & 5.5M  & 375M & 74.9 & - \\
        ProxylessNAS \cite{proxyless19} & auto    & 4.1M  & 320M & 74.6 & 92.2 \\
        ProxylessNAS (1.4x) & auto    & 6.9M  & 581M & 76.7 & 93.3 \\  
        MobileNetV3-Large \cite{mobilenetv319} & combined    & 5.4M  & 217M & 75.2 & - \\  
        MobileNetV3-Large (1.25x) &  combined  & 7.5M  & 356M & 76.6 & - \\  
        \midrule[0.1em]
        \bf MixNet-S    &  \bf auto    & \bf  4.1M  & \bf 256M & \bf 75.8 & \bf 92.8 \\
		\bf MixNet-M    &  \bf auto    &  \bf 5.0M  & \bf 360M & \bf 77.0 & \bf 93.3  \\  
		\bf MixNet-L    &  \bf auto    &  \bf 7.3M  & \bf 565M & \bf 78.9 & \bf 94.2  \\          \toprule[0.2em]                                                             
        \end{tabular}
    } 
    \vskip 0.05in
    \caption{
        \textbf{MixNet performance results on ImageNet 2012~\cite{imagenet15}}.
    }
    \vskip -0.1in                                           
    \label{tab:imagenet}       
\end{table}      

\begin{itemize}
	\itemsep0em 
	\item \BF{3x3}: MixConv with one group of filters ($g=1$) with kernel size 3x3.
	\item ...
	\item \BF{3x3, 5x5, 7x7, 9x9, 11x11}: MixConv with five groups of filters ($g=5$) with kernel size \{3x3, 5x5, 7x7, 9x9, 11x11\}. Each group has roughly the same number of channels. 
\end{itemize}

\noindent In order to simplify the search process, we don't include exponential channel partition or dilated convolutions in our search space, but it is trivial to integrate them in future work.

Similar to recent neural architecture search approaches \cite{mnas19,proxyless19,fbnet19}, we directly search on ImageNet train set, and then pick a few top-performing models from search to verify their accuracy on ImageNet validation set and transfer learning datasets.

\subsection{MixNet Performance on ImageNet}

Table \ref{tab:imagenet} shows the ImageNet performance of MixNets. Here we obtain MixNet-S and M from neural architecture search, and scale up MixNet-M with depth multiplier 1.3 to obtain MixNet-L. All models are trained with the same settings as MnasNet \cite{mnas19}.

In general, our MixNets outperform all latest mobile ConvNets: Compared to the hand-crafted models, our MixNets improve top-1 accuracy by 4.2\% than MobileNetV2 \cite{mobilenetv218} and 3.5\% than ShuffleNetV2 \cite{shufflenetv218}, under the same FLOPS constraint;  Compared to the latest automated models, our MixNets  achieve better accuracy than MnasNet (+1.3\%), FBNets (+2.0\%), ProxylessNAS (+2.2\%) under similar FLOPS constraint. Our models also achieve similar performance as the latest MobileNetV3 \cite{mobilenetv319}, which is developed concurrently with our work with several manual optimizations in addition of architecture search. In particular, our MixNet-L achieves a new state-of-the-art 78.9\% top-1 accuracy under typical mobile FLOPS (<600M) constraint.
\begin{figure}[t]
	\centering 
	\includegraphics[width=0.75\linewidth,keepaspectratio=true]{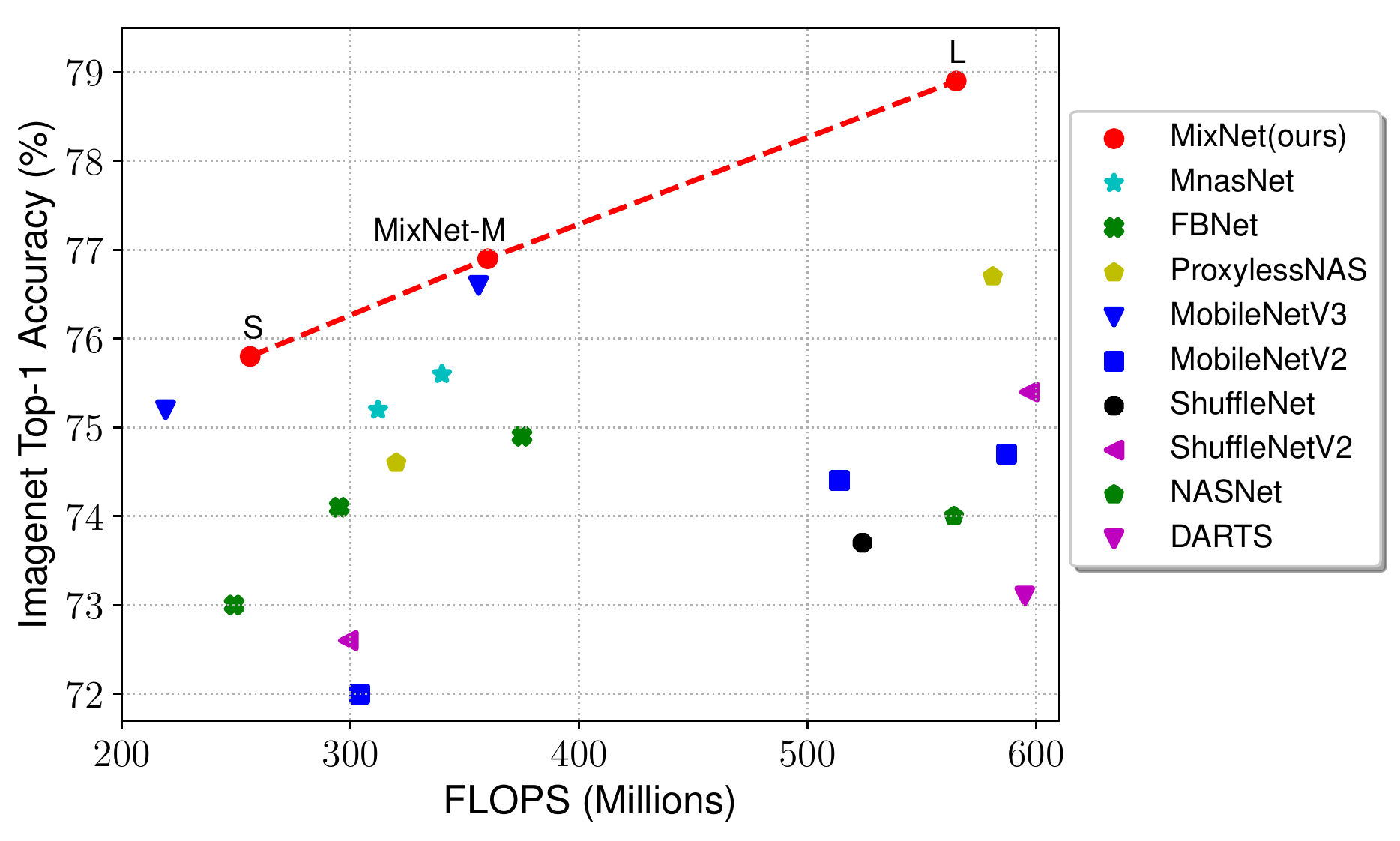}
	\vskip -0.1in
	\caption{
		\textbf{ImageNet performance comparison.}
	}
	\label{fig:nasflopsparams}
\end{figure} %
\begin{figure}                                           
	\centering                                                                  
	\includegraphics[width=\linewidth,keepaspectratio=true]{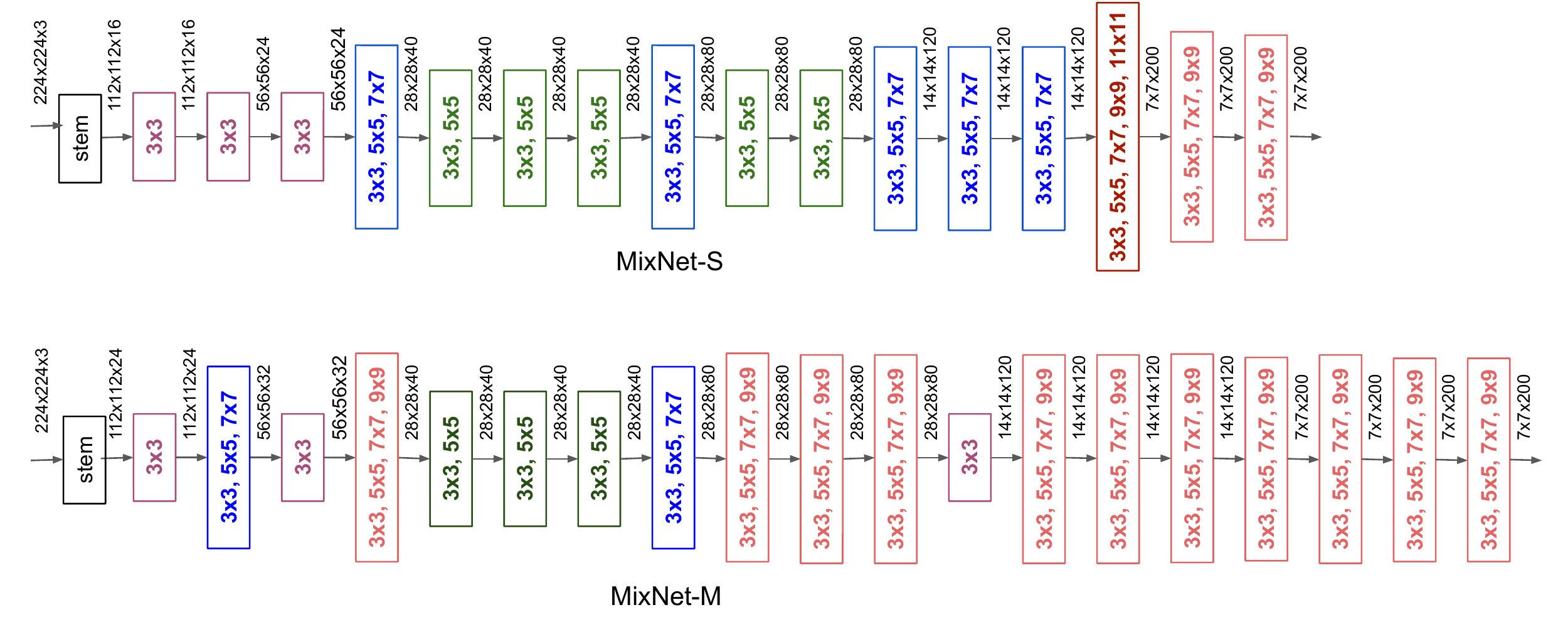}
	\vskip -0.1in
	\caption{
		\textbf{MixNet architectures} --  \TT{MixNet-S} and \TT{MixNet-M} are from Table \ref{tab:imagenet}. We mainly highlight MixConv kernel size (e.g. \{3x3, 5x5\}) and input/output tensor shape.
	}
	\label{fig:nasarch}
\end{figure} 
Figure \ref{fig:nasflopsparams} visualizes the ImageNet performance comparison. We observe that recent progresses on neural architecture search have significantly improved model performance \cite{mnas19,fbnet19,proxyless19} than previous hand-crafted mobile ConvNets \cite{mobilenetv218,shufflenetv218}. However, by introducing a new type of efficient MixConv, we can further improve model accuracy and efficiency based on the same neural architecture search techniques.

\subsection{MixNet Architectures}

To understand why our MixNets achieve better accuracy and efficiency, Figure \ref{fig:nasarch} illustrates the network architecture for \TT{MixNet-S} and \TT{MixNet-M} from Table \ref{tab:imagenet}. In general, they both use a variety of MixConv with different kernel sizes throughout the network: small kernels are more common in early stage for saving computational cost, while large kernels are more common in later stage for better accuracy. We also observe that the bigger \TT{MixNet-M} tends to use more large kernels and more layers to pursing higher accuracy, with the cost of more parameters and FLOPS. Unlike vanilla depthwise convolutions that suffer from serious accuracy degradation for large kernel sizes (Figure \ref{fig:limits}), our MixNets are capable of utilizing very large kernels such as 9x9 and 11x11 to capture high-resolution patterns from input images,  without hurting model accuracy and efficiency.

\subsection{Transfer Learning Performance}

We have also evaluated our MixNets on four widely used transfer learning datasets, including CIFAR-10/100 \cite{cifar}, Oxford-IIIT Pets \cite{oxfordpets} , and Food-101 \cite{food101}. Table \ref{tab:transdataset} shows their statistics of train set size, test set size, and number of classes.

\begin{figure}                                           
	\centering                                                                  
	\includegraphics[width=\linewidth,keepaspectratio=true]{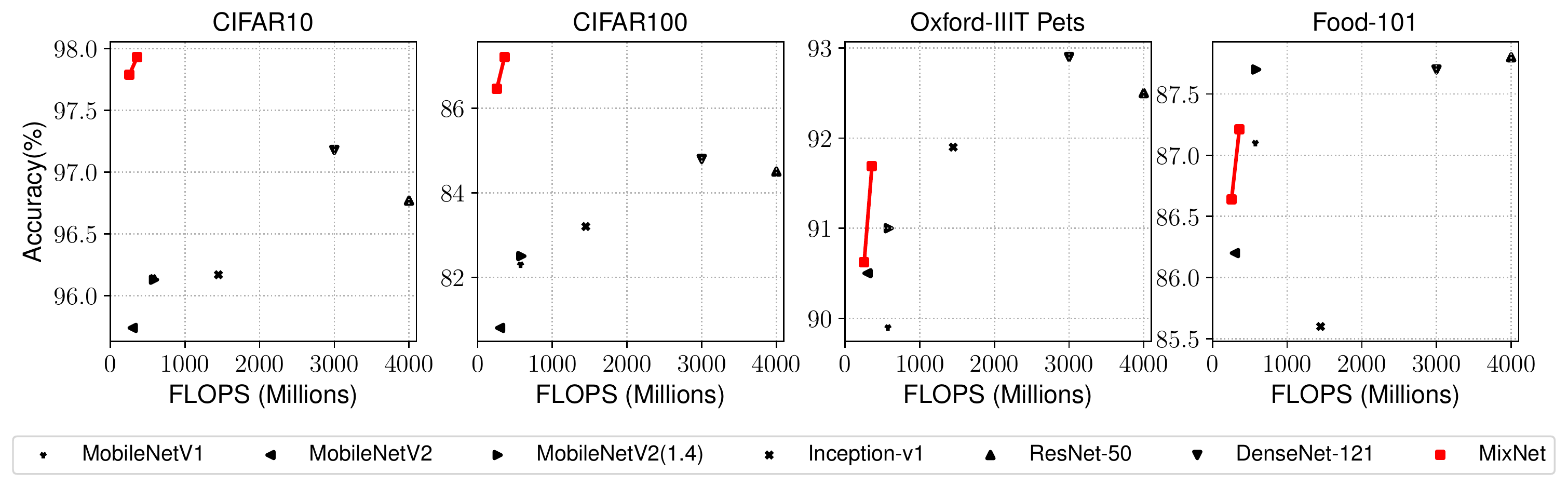}
	\vskip -0.1in
	\caption{
		\textbf{Transfer learning performance} -- \TT{MixNet-S/M} are from Table \ref{tab:imagenet}.
	}
	\label{fig:nastransfer}
\end{figure}

Figure \ref{fig:nastransfer} compares our \TT{MixNet-S/M} with a list of previous models on transfer learning accuracy and  FLOPS. For each model, we first train it from scratch on ImageNet and than finetune all the weights on the target dataset using similar settings as \cite{imagenettransfer18}. The accuracy and FLOPS data for MobileNets \cite{mobilenetv117,mobilenetv218}, Inception \cite{googlenet14}, ResNet \cite{resnet16}, DenseNet \cite{densenet17} are from \cite{imagenettransfer18}. In general, our MixNets significantly outperform previous models on all these datasets, especially on the most widely used CIFAR-10 and CIFAR-100, suggesting our MixNets also generalize well to transfer learning.
In particular, our \TT{MixNet-M} achieves 97.92\% accuracy with 3.49M parameters and 352M FLOPS,  which is \BF{11.4x} more efficient with 1\% higher accuracy than ResNet-50  \cite{resnet16}.

\begin{table}                                            
  \centering                                                                        
  \resizebox{0.55\linewidth}{!}{                                                   
        \begin{tabular}{c|ccc}                                                      
                \toprule[0.2em]                                                     
         Dataset      & TrainSize  &   TestSize & Classes \\                    
         \midrule[0.1em]                                                            
          CIFAR-10 \cite{cifar} &  50,000 & 10,000 & 10 \\                          
          CIFAR-100 \cite{cifar} &  50,000 & 10,000 & 100  \\                       
        Oxford-IIIT Pets \cite{oxfordpets} & 3,680 & 3,369 & 37 \\          
        Food-101 \cite{food101} & 75,750 & 25,250 & 101   \\                
         \bottomrule[0.2em]                                                          
    \end{tabular}                                                                   
  }                    
  \vskip 0.1in
  \caption{                                                                         
	\small \textbf{Transfer learning datasets.}
  }                                                          
  \label{tab:transdataset}             
\end{table}    %

\section{Conclusions}
\label{sec:conclude}

In this paper, we revisit the impact of kernel size for depthwise convolution, and identify that traditional depthwise convolution suffers from the limitations of single kernel size. To address this issue, we  proposes MixConv, which mixes multiple kernels in a single op to take advantage of different kernel sizes. We show that our MixConv is a simple drop-in replacement of vanilla depthwise convolution, and improves the accuracy and efficiency for MobileNets, on both image classification and object detection tasks. Based on our proposed MixConv, we further develop a new family of MixNets using neural architecture search techniques. Experimental results show that our MixNets achieve significantly better accuracy and efficiency than all latest mobile ConvNets on both ImageNet classification and four widely used transfer learning datasets. 
\bibliography{cv}
\end{document}